# Generating Negations of Probability Distributions

Ildar Batyrshin[1][0000-0003-0241-7902], Luis Alfonso Villa-Vargas[1], Marco Antonio Ramírez-Salinas[1], Moisés Salinas-Rosales[1], Nailya Kubysheva[2][0000-0002-5582-5814]

[1]Instituto Politécnico Nacional, Centro de Investigación en Computación, CDMX, México
batyr1@gmail.com
[2]Kazan Federal University
aibolit70@mail.ru

**Abstract.** Recently it was introduced a negation of a probability distribution. The need for such negation arises when a knowledge-based system can use the terms like *NOT HIGH,* where *HIGH* is represented by a probability distribution (pd). For example, *HIGH PROFIT* or *HIGH PRICE* can be considered. The application of this negation in Dempster-Shafer theory was considered in many works. Although several negations of probability distributions have been proposed, it was not clear how to construct other negation. In this paper, we consider negations of probability distributions as point-by-point transformations of pd using decreasing functions defined on [0,1] called negators. We propose the general method of generation of negators and corresponding negations of pd, and study their properties. We give a characterization of linear negators as a convex combination of Yager's and uniform negators.

## 1. Introduction

The concept of negation of probability distribution (pd) was recently introduced by Yager [18]. He defined the negation $\bar{P}$ of a finite probability distribution $P = (p_1, ..., p_n)$ by: $\bar{P} = (\overline{p_1}, ..., \overline{p_n})$, where $\overline{p_i}$ is defined by: $\overline{p_i} = \frac{1-p_i}{n-1}$. He noted that the last function is decreasing, and the negation of probability distribution (pd) increases its entropy. Yager proposed that the negation of pd can be used in a knowledge-based system operating with concepts like NOT HIGH, where HIGH is represented by a probability distribution. Similar concepts like HIGH PROFIT or HIGH PRICE can appear in soft computing models with economic applications [2, 7, 11]. Negations of probability distributions and their application in Dempster-Shafer theory was considered in many works [4-6, 8, 9, 12-20]. Yager mentioned [18] that there could exist other negations of probability distributions (pd). In [20], it was introduced another example of the negation of probability distributions based on Tsallis entropy. But how to construct other negations of pd is still not clear. Moreover, the general properties of the negations of pd were net studied.

In this paper, we study a special type of point-by-point transformations and negations of probability distributions based on transformation functions and negators. Negators are introduced as decreasing functions defined on the set of probability values [0,1]. Although some properties of negators may be related to the properties of fuzzy negations, negators differ from negation operations of fuzzy logic [1, 15]. We propose the general method of generation of negators and study their properties. We introduce the parametric class of linear negations, including Yagers's negation as a particular case. We give a characterization of such negators and show that they can be represented as a convex combination of Yager's and uniform negators.

The paper has the following structure. In Section 2, we consider transformations of probability distributions that can be used as negations of probability distributions. In Section 3, we consider the functions defined on [0,1], called transformation functions or negators, that can be used for point-by-point transformation of a probability distribution into its negation. We propose a method of construction of negators using generator functions and using a weighted sum of other negators. Section 4 studies the properties of negators and negations of pd related to the fixed points of these transformations. Section 5 studies the properties of images of pd-independent transformation functions and negators. Section 6 introduces the parametric class of linear negators as a convex combination of Yager's and uniform negators. Section 7 contains conclusions.

## 2. Transformations and Negations of Probability Distributions

A sequence $P = (p_1, \ldots, p_n)$ of $n$ real values $p_i$ satisfying for all $i = 1, \ldots, n$, $(n \geq 2)$, the properties:

$$0 \leq p_i \leq 1, \quad \sum_{i=1}^{n} p_i = 1, \tag{1}$$

will be called a *probability distribution* (*pd*) *of the length n*. One can consider $p_i$ as a probability of an event $x_i$ in some experiment $X$ with outcomes $\{x_1, \ldots, x_n\}$, $n \geq 2$. All probability distributions discussed in this paper will have the same length $n$.

Consider the simplest examples of probability distributions.

The probability distribution $P_U = (p_1, \ldots, p_n)$ will be called the *uniform distribution* if:

$$p_i = p_j \text{ for all } i, j = 1, \ldots, n. \tag{2}$$

From (2) and (1) we have:

$$p_i = \frac{1}{n} \text{ for all } i = 1, \ldots, n, \quad \text{and} \quad P_U = \left(\frac{1}{n}, \ldots, \frac{1}{n}\right). \tag{3}$$

The probability distribution $P_{(i)} = (p_1, \ldots, p_n)$ satisfying the property: $p_i = 1$ for some $i$ in $\{1, \ldots, n\}$, and $p_j = 0$ for all $j \neq i$, will be referred to as a *degenerate* or *point distribution* [3]. For example, for $i = 1$ and $i = n$ we have the following point distributions: $P_{(1)} = (1, 0, \ldots, 0)$, and $P_{(n)} = (0, \ldots, 0, 1)$.

Let $\mathcal{P}_n$ be the set of all probability distributions of the length $n$. In this paper, we will consider *transformations of probability distributions* $P = (p_1, \ldots, p_n)$ into probability distributions $TR(P) = Q = (q_1, \ldots, q_n)$ satisfying for all $i, j = 1, \ldots, n$, the following properties:

$$0 \leq q_i \leq 1, \sum_{i=1}^{n} q_i = 1, \tag{4}$$

$$\text{if } p_i = p_j \text{ then } q_i = q_j. \tag{5}$$

A transformation of probability distributions will be called a *negation of probability distributions* and denoted as $neg(P)$ if it transforms any pd $P = (p_1, \ldots, p_n)$ into a probability distribution $neg(P) = Q = (q_1, \ldots, q_n)$ satisfying for all $i, j = 1, \ldots, n$, the following property:

$$\text{if } p_i \leq p_j, \text{ then } q_i \geq q_j. \tag{6}$$

Note that from (6), it follows (5).

In this paper, we will study a special type of point-by-point transformations and negations of probability distributions based on transformation functions and negators considered below.

### 3. Transformation Functions and Negators

A *transformation function* is a function of probability values $N(p)$ transforming point-by-point any probability distributions $P = (p_1, \ldots, p_n)$ into a probability distribution: $TR_N(P) = (N(p_1), \ldots, N(p_n))$, such that for all $i = 1, \ldots, n$, it is fulfilled:

$$0 \leq N(p_i) \leq 1, \sum_{i=1}^{n} N(p_i) = 1, \tag{7}$$

$$\text{if } p_i = p_j \text{ then } N(p_i) = N(p_j). \tag{8}$$

A transformation function $N$ will be called *pd-independent* or *independent* if for any probability distributions $P = (p_1, \ldots, p_n)$ and $Q = (q_1, \ldots, q_n)$ for all $i, j = 1, \ldots, n$ it is fulfilled:

$$\text{if } p_i = q_j, \text{ then } N(p_i) = N(q_j). \tag{9}$$

A transformation function $N$ will be called a *negator* if it is a decreasing function, such that for all probability distributions $P = (p_1, \ldots, p_n)$ and all $i, j = 1, \ldots, n$ it is fulfilled:

$$\text{if } p_i \leq p_j, \text{ then } N(p_i) \geq N(p_j). \tag{10}$$

A negator $N$ transforms point-by-point the probability distribution $P = (p_1, \ldots, p_n)$ into its negation: $neg_N(P) = (N(p_1), \ldots, N(p_n))$. A pd-independent negator $N$ is a decreasing function independent on any probability distribution and satisfying the properties (7), (9) and (10), see Example 1.

**Theorem 1.** Let $P = (p_1, \ldots, p_n)$ be an arbitrary probability distribution and $f(p)$ be a non-negative real-valued function satisfying the property: $\sum_{i=1}^{n} f(p_i) > 0$, then the function $N$ defined for all $i = 1, \ldots, n$ by:

$$N(p_i) = \frac{f(p_i)}{\sum_{i=1}^{n} f(p_i)}, \tag{11}$$

is a transformation function. If $f(p)$ is a decreasing function of $p$, then $N$ is a negator.

The proof is straightforward taking into account that $\sum_{i=1}^{n} N(p_i) = \sum_{i=1}^{n} \frac{f(p_i)}{\sum_{i=1}^{n} f(p_i)} = 1$.

The function $f$ in (11) will be called a *generator* of a transformation function or negator $N$.

**Example 1.** Consider transformation functions and negators $N$ with corresponding generators defined for all $P = (p_1, \ldots, p_n)$ and for all $i = 1, \ldots, n$ as follows:

$$N_I(p_i) = p_i, \qquad f(p_i) = p_i,$$

$$N_{RS}(p_i) = \frac{\sqrt{p_i}}{\sum_{i=1}^{n} \sqrt{p_i}}, \qquad f(p_i) = \sqrt{p_i},$$

$$N_U(p_i) = \frac{1}{n}, \qquad f(p_i) = 1,$$

$$N_Y(p_i) = \frac{1-p_i}{n-1}, \qquad f(p_i) = 1 - p_i,$$

$$N_T(p_i) = \frac{1-p_i^k}{n-\sum_{i=1}^n p_i^k}, \qquad f(p_i) = 1 - p_i^k, \ k \neq 0.$$

The function $N_I$ defines the *identity transformation* $TR_{N_I}(P)$ of any probability distribution into itself: $TR_{N_I}(P) = (N_I(p_1), \ldots, N_I(p_n)) = (p_1, \ldots, p_n) = P$. The function $N_I$ is strictly increasing hence it is not a negator and the transformation $TR_{N_I}$ is not a negation of probability distributions.

The function $N_{RS}$ is the strictly increasing transformation function, and hence it is not a negator.

The function $N_U$ is decreasing (non-increasing). It will be referred to as the *uniform negator*. It ignores information about the probability values in $P$ and defines the transformation of any pd into the uniform distribution: $neg_{N_U}(P) = (N_U(p_1), \ldots, N_U(p_n)) = \left(\frac{1}{n}, \ldots, \frac{1}{n}\right) = P_U$.

The negator $N_Y$ is introduced by Yager [18]. This negator defines the following negation of a probability distribution $P$: $neg_{N_Y}(P) = (N_Y(p_1), \ldots, N_Y(p_n)) = \left(\frac{1-p_1}{n-1}, \ldots, \frac{1-p_n}{n-1}\right)$. For example, for the point distribution $P_{(1)} = (1,0,\ldots,0)$ we obtain: $neg_{N_Y}((1,0,\ldots,0)) = \left(0, \frac{1}{n-1}, \ldots, \frac{1}{n-1}\right)$.

The negator $N_T$ is based on Tsallis entropy [20].

The transformation function $N_{RS}$ and negator $N_T$ generally depend on probability distributions. For example, the property (9) will not be fulfilled for the negator $N_T$ for $k \neq 1$ for some pd $P = (p_1, \ldots, p_n)$ and $Q = (q_1, \ldots, q_n)$ if $\sum_{i=1}^n p_i^k \neq \sum_{i=1}^n q_i^k$.

The transformation function $N_I$ and negators $N_U$ and $N_Y$ are pd-independent.

**Theorem 2.** Let $N_1, \ldots, N_m$ be transformation functions, $w_1, \ldots, w_m$, are the weights, such that $0 \leq w_k \leq 1$ for all $k = 1, \ldots, m$, and $\sum_{k=1}^m w_k = 1$, then

$$N(p_i) = \sum_{k=1}^m w_k \cdot N_k(p_i), \ i = 1, \ldots, n, \tag{12}$$

is a transformation function. If all $N_1, \ldots, N_m$ are negators then $N$ is a negator. If all $N_k$ are pd-independent then $N$ is pd-independent.

**Proof.** From the definition of $N(p_i)$ it follows that $N(p_i) \geq 0, i = 1, \ldots, n$, and together with

$$\sum_{i=1}^n N(p_i) = \sum_{i=1}^n \sum_{k=1}^m w_k N_k(p_i) = \sum_{k=1}^m \sum_{i=1}^n w_k N_k(p_i) = \sum_{k=1}^m w_k \sum_{i=1}^n N_k(p_i) = \sum_{k=1}^m w_k \cdot 1 = 1$$

we obtain $0 \leq N(p_i) \leq 1$, i.e., (7) is fulfilled, and $N$ is a transformation function. If $p_i \leq p_j$, then $N_k(p_i) \geq N_k(p_j)$ for all $k = 1, \ldots, m$, and from (12) it follows $N(p_i) \geq N(p_j)$, i.e., $N$ is decreasing. Hence $N$ is a negator. It is clear that $N$ is pd-independent if all negators $N_k$ are pd-independent ∎

## 4. Fixed points of Transformations of Probability Distributions and Negators

A probability distribution $P$ will be called a *fixed point* of the transformation of probability distributions $TR(P)$ if

$$TR(P) = P.$$

An element $p$ in $[0,1]$ will be called a *fixed point of* a transformation function $N$ if

$$N(p) = p.$$

**Proposition 1.** The universal distribution $P_U = \left(\frac{1}{n}, \ldots, \frac{1}{n}\right)$ is a fixed point of any transformation $TR$ of probability distributions, i.e.:

$$TR(P_U) = P_U. \tag{13}$$

**Proof.** For any transformation $TR$ of $P_U = (p_1, \ldots, p_n) = \left(\frac{1}{n}, \ldots, \frac{1}{n}\right)$, we have: $TR(P_U) = TR((p_1, \ldots, p_n)) = TR\left(\left(\frac{1}{n}, \ldots, \frac{1}{n}\right)\right) = (q_1, \ldots, q_n)$, where from (5) and (4) it follows: $q_i = q_j = \frac{1}{n}$, for all $i, j = 1, \ldots, n$, hence $TR(P_U) = P_U$ ∎

For the pd-independent transformation function $N_I$ from Example 1, any element $p$ in $[0,1]$ will be a fixed point. Also, any probability distribution $P$ will be a fixed point of the identity transformation: $TR_{N_I}(P) = (N_I(p_1), \ldots, N_I(p_n)) = (p_1, \ldots, p_n) = P$. One can check that the probability distribution $P_U = \left(\frac{1}{n}, \ldots, \frac{1}{n}\right)$ will be the fixed point of all transformations and negations of probability distributions defined by transformation functions from Example 1.

**Proposition 2.** Any pd-independent transformation function $N$ has the fixed point $p = \frac{1}{n}$, i.e.:

$$N\left(\frac{1}{n}\right) = \frac{1}{n}. \tag{14}$$

**Proof.** From the definition of the pd-independent transformation functions and from Proposition 1 it follows: $TR_N\left(\left(\frac{1}{n}, \ldots, \frac{1}{n}\right)\right) = \left(N\left(\frac{1}{n}\right), \ldots, N\left(\frac{1}{n}\right)\right) = \left(\frac{1}{n}, \ldots, \frac{1}{n}\right)$, i.e., $N\left(\frac{1}{n}\right) = \frac{1}{n}$ ∎

**Theorem 3.** Any pd-independent negator $N$ has the unique fixed point $p = \frac{1}{n}$, i.e., $N\left(\frac{1}{n}\right) = \frac{1}{n}$, and any negation of probability distributions $neg_N$ has a unique fixed point $P_U$.

**Proof.** From Propositions 1 and 2 it follows that $p = \frac{1}{n}$ is a fixed point of any pd-independent negator $N$, and $P_U$ is the fixed point of any negation $neg_N$. We need to prove the uniqueness of these fixed points. Suppose that $q \in [0,1]$ is a different from $\frac{1}{n}$ fixed point of some pd-independent negator $N$, i.e., $N(q) = q$, and $q \neq \frac{1}{n}$. If $q < \frac{1}{n}$ then from (10) and (14), it follows: $N(q) \geq N\left(\frac{1}{n}\right) = \frac{1}{n} > q$, that contradicts to the assumption $N(q) = q$. The similar contradiction we obtain if $q > \frac{1}{n}$. Hence $p = \frac{1}{n}$ is the unique fixed point of $N$.

Let $P = (p_1, \ldots, p_n)$ be a fixed point of $neg_N$ for some independent negator $N$. Then $neg_N(P) = (N(p_1), \ldots, N(p_n)) = P = (p_1, \ldots, p_n)$, and $N(p_i) = p_i$ for all $i = 1, \ldots, n$. Since $\frac{1}{n}$ is the unique fixed point of any independent negator $N$, we obtain $p_i = \frac{1}{n}$ for all $i = 1, \ldots, n$, hence $P = (p_1, \ldots, p_n) = \left(\frac{1}{n}, \ldots, \frac{1}{n}\right) = P_U$, i.e. $P_U$ is the unique fixed point of any negation of probability distributions $neg_N$ generated by a pd-independent negator $N$ ∎

Below we will describe the images of negators to the left and right of the fixed point $N\left(\frac{1}{n}\right) = \frac{1}{n}$.

## 5. The Properties of Images of pd-Independent Negators

**Proposition 3.** Let $N$ be an independent transformation function then for any $p$ in [0,1] it is fulfilled:

$$N\left(\frac{1-p}{n-1}\right) = \frac{1-N(p)}{n-1}. \tag{15}$$

**Proof.** Let us show that (15) is fulfilled for any $p$ in [0,1]. Consider the probability distribution $P = (p_1, \ldots, p_n)$ such that $p_1 = p$ and $p_i = p_j$ for all $i, j \geq 2$. Denote $q = p_i$ for $i \geq 2$. Then $P = (p_1, \ldots, p_n) = (p, q, \ldots, q)$, and from $\sum_{i=1}^n p_i = 1$, we obtain:

$$q = \frac{1-p}{n-1}. \tag{16}$$

For the probability distribution $TR_N(P) = (N(p_1), \ldots, N(p_n)) = (N(p), N(q), \ldots, N(q))$, from $\sum_{i=1}^n N(p_i) = 1$, it follows: $N(q) = \frac{1-N(p)}{n-1}$, and from (16) we obtain: $N\left(\frac{1-p}{n-1}\right) = \frac{1-N(p)}{n-1}$ ∎

**Proposition 4.** For independent transformation function $N$ it is fulfilled:

$$N(0) = \frac{1-N(1)}{n-1}. \tag{17}$$

**Proof.** From (15) we obtain for $p = 1$: $N\left(\frac{1-1}{n-1}\right) = N(0) = \frac{1-N(1)}{n-1}$ ∎

Below we consider the properties of independent negators following from the properties (15) and (17).

**Proposition 5.** For any independent negator $N$ it is fulfilled:

$$N(1) \in \left[0, \frac{1}{n}\right], \tag{18}$$

$$N(0) \in \left[\frac{1}{n}, \frac{1}{n-1}\right]. \tag{19}$$

**Proof.** From $1 > \frac{1}{n}$, (10) and (14) it follows: $N(1) \leq N\left(\frac{1}{n}\right) = \frac{1}{n}$, and hence (18) is fulfilled. From (17) we obtain for border cases of (18): for $N(1) = 0$, $N(0) = \frac{1-0}{n-1} = \frac{1}{n-1}$, and for $N(1) = \frac{1}{n}$, $N(0) = \frac{1-\frac{1}{n}}{n-1} = \frac{1}{n}$, that together with (18) and (10) gives (19) ∎

The following Theorem extends the results of the previous propositions.

**Theorem 4.** For any independent negator $N$ it is fulfilled:

$$N(p) \in [0, \tfrac{1}{n}] \quad \text{if } p \geq \tfrac{1}{n}, \tag{20}$$

$$N(p) \in [\tfrac{1}{n}, \tfrac{1}{n-1}] \quad \text{if } p \leq \tfrac{1}{n}. \tag{21}$$

**Proof.** Since $N$ is a non-negative decreasing function and independent negator from (10) and from $N\left(\tfrac{1}{n}\right) = \tfrac{1}{n}$ we obtain:

if $\tfrac{1}{n} \leq p$ then $N(p) \leq N\left(\tfrac{1}{n}\right) = \tfrac{1}{n}$, and (20) is fulfilled,

if $0 \leq p \leq \tfrac{1}{n}$ then $N(0) \geq N(p) \geq N\left(\tfrac{1}{n}\right) = \tfrac{1}{n}$, and from (19) it follows (21) ∎

From Theorem 3 it follows that $p = \tfrac{1}{n}$ is the unique fixed point of any independent negator $N$, i.e., $N\left(\tfrac{1}{n}\right) = \tfrac{1}{n}$, and Theorem 4 describes the values of such negators to the left and right of this fixed point.

Using the results of this Section, in the following Section, we characterize linear independent negators.

## 6. Parametric class of linear negators

We will say that an independent negator $N$ is *linear* if $N(p)$ is a linear function of $p \in [0,1]$. The negation $neg_N(P) = (N(p_1), \dots, N(p_n))$ of a probability distribution $P = (p_1, \dots, p_n)$ will be called a *linear negation* of pd if $N(p)$ is a linear negator.

Let us consider the convex combination of negators $N_U$ and $N_Y$:

$$N(p) = \alpha N_U(p) + (1-\alpha) N_Y(p) = \alpha \tfrac{1}{n} + (1-\alpha)\tfrac{1-p}{n-1}, \tag{22}$$

where $\alpha \in [0,1]$ is a parameter of convex combination, and $p$ is a probability. Since $N_U$ and $N_Y$ are pd-independent negators from Theorem 2 it follows that (22) is also an independent negator. From (18) we have $N(1) \in \left[0, \tfrac{1}{n}\right]$ and $nN(1) \in [0,1]$, hence in (22), we can use $\alpha = nN(1)$, and represent (22) as follows:

$$N(p) = N(1) + \left(1 - nN(1)\right)\tfrac{1-p}{n-1}. \tag{23}$$

For $p = 0$ from (23) we obtain (compare also with (17)): $N(0) = \tfrac{1-N(1)}{n-1}$, and then:

$$N(1) = 1 - (n-1)N(0). \tag{24}$$

Replacing $N(1)$ in (23) by (24) after equivalent transformations we can represent (23) as follows:

$$N(p) = N(0) + \left(1 - nN(0)\right)p. \tag{25}$$

**Theorem 5.** A function $N(p)$ is a linear negator if and only if it is a convex combination of negators $N_U$ and $N_Y$, i.e. for some $\alpha \in [0,1]$ for all $p$ in $[0,1]$ it is fulfilled (22).

**Proof.** From Theorem 2 it follows that (22) is a linear negator. Let us show that any linear negator $N^*$ can be represented by (22) for some $\alpha \in [0,1]$. Due to (14) and (18) the line corresponding to the linear function $N^*$ passes through two points $(p, N^*(p))$ with coordinates $p_1 = \frac{1}{n}$, $N^*(p_1) = N^*\left(\frac{1}{n}\right) = \frac{1}{n}$, and $p_2 = 1, N^*(p_2) = N^*(1)$, where $N^*(1) \in \left[0, \frac{1}{n}\right]$, see Fig. 1. Since only one line passes through two different points, it is sufficient to show that for some $\alpha$ the line corresponding to the linear function (22) also passes through these two points. Since (14) fulfills for all independent negators then it is sufficient to show that for some $\alpha$ the linear negator (22) have the value $N(1) = N^*(1)$. Denote $\alpha = nN^*(1)$, and obtain from (22) and (23) the linear negator: $N(p) = N^*(1) + (1 - nN^*(1))\frac{1-p}{n-1}$, such that $N(1) = N^*(1)$ ∎

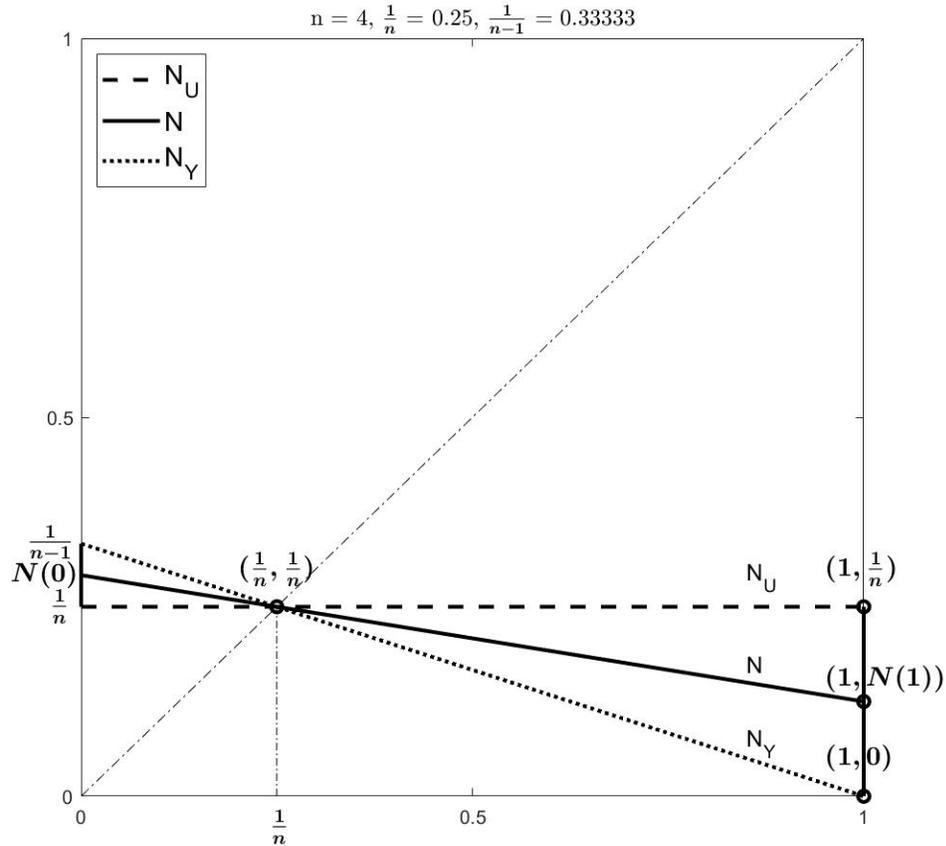

**Figure 1.** Linear negator $N$ defined by (22) for $n = 4$ as a convex combination of negators $N_U$ and $N_Y$.

The Yager's negator $N_Y$ and the uniform negator $N_U$ are the border cases of linear negators (22). When $N(1) = 0$ we obtain from (23): $N(p) = \frac{1-p}{n-1} = N_Y$, and when $N(1) = \frac{1}{n}$ we obtain from (23): $N(p) = N(1) = \frac{1}{n} = N_U$.

**Example 2.** Consider probability distribution $P = (0, 0.1, 0.2, 0.3, 0.4)$. We have $n = 5$ and fixed point $\frac{1}{n} = \frac{1}{5} = 0.2$. For any negator we have $N(0.2) = 0.2$. From (23) we have: $N(p) = N(1) +$

$(1 - nN(1))\frac{1-p}{n-1} = N(1) + (1 - 5N(1))\frac{1-p}{4} = \frac{1}{4}[1 - N(1) - (1 - 5N(1))p]$. Construct linear negator for $N(1) = \frac{1}{2n} = \frac{1}{10} = 0.1$. Then we obtain: $N(p) = \frac{1}{4}[1 - 0.1 - 0.5p] = 0.25(0.9 - 0.5p)$. Calculating negators for all components of pd $P = (0, 0.1, 0.2, 0.3, 0.4)$ we obtain: $neg_N(P) = (0.225, 0.2125, 0.2, 0.1875, 0.175)$. Compare with Yager's negation of $P$: $neg_{N_Y}(P) = (0.25, 0.225, 0.2, 0.175, 0.15)$ and with uniform negation $neg_{N_U}(P) = (0.2, 0.2, 0.2, 0.2, 0.2)$. As we can see, the negation $neg_N(P)$ is closer to $neg_{N_U}$ than $neg_{N_Y}(P)$.

In [18], it was considered the entropy of probability distribution defined as follows:

$$H(P) = \sum_{i=1}^{n}(1 - p_i)p_i.$$

It was shown [18] that Yager's negation of pd (based on Yager's negator) increases the entropy of probability distribution. The uniform distribution has the maximal entropy, hence the negation of pd defined by the uniform negator $N_U$ gives the maximal increase in the entropy, comparing with other possible negations of probability distributions. As a result, linear negators constructed in (22) as a convex combination of Yager's negator and the uniform negators will define linear negation of pd increasing the entropy of probability distributions.

## 7. Conclusion

The paper studied the general properties of functions defined on [0,1], called negators, that can be used for a point-by-point transformation of probability distributions into negations of these distributions. We proposed a method of construction of negators using generator functions and using a weighted sum of other negators. The paper studied the properties of negators and negations of pd related to the fixed points of these transformations. We considered two classes on negators: dependent and independent on probability distributions. We introduced the concept of linear negators as decreasing linear functions defined on [0,1] and independent on probability distributions. These linear negators are characterized as convex combinations of Yager's and uniform negators.

As future work, it is supposed to solve the Open Problem: prove or disprove a hypothesis that any independent negator is linear. Also, it is supposed to develop probabilistic models of decision making in economics based on negations of probability distributions considering some approaches studied in [2, 7, 11].

**Acknowledgments.** The research is partially supported by the project IPN SIP 20200853.

**References**

1. Batyrshin, I. (2003). On the structure of involutive, contracting and expanding negations. Fuzzy Sets and Systems, 139(3), 661-672.
2. Dymowa, L. (2011). Soft computing in economics and finance. Heidelberg: Springer.
3. Evans, M. J., & Rosenthal, J. S. (2010). Probability and statistics: The science of uncertainty. 2nd ed. W. H. Freeman and Company, New York.
4. Gao, X., & Deng, Y. (2019). The negation of basic probability assignment. IEEE Access, 7, 107006-107014.
5. Gao, X., & Deng, Y. (2019). The generalization negation of probability distribution and its application in target recognition based on sensor fusion. International Journal of Distributed Sensor Networks, 15(5), 1550147719849381.

6. Gao, X., & Deng, Y. (2020). Quantum model of mass function. International Journal of Intelligent Systems, 35(2), 267-282.
7. Kreinovich, V., Thach, N. N., Trung, N. D., & Van Thanh, D. (Eds.). (2019). Beyond Traditional Probabilistic Methods in Economics (Vol. 809). Springer.
8. Li, S., Xiao, F., & Abawajy, J. H. (2020). Conflict Management of evidence theory based on belief entropy and negation. IEEE Access, 8, 37766-37774.
9. Luo, Z., & Deng, Y. (2019). A matrix method of basic belief assignment's negation in Dempster-Shafer theory. IEEE Transactions on Fuzzy Systems.
10. Rockafellar, R. Tyrrell (1970). Convex Analysis, Princeton Mathematical Series, 28, Princeton University Press, Princeton, N.J.
11. Sriboonchita, S., Wong, W. K., Dhompongsa, S., & Nguyen, H. T. (2009). Stochastic dominance and applications to finance, risk and economics. CRC Press.
12. Srivastava, A., & Kaur, L. (2019). Uncertainty and negation—Information theoretic applications. International Journal of Intelligent Systems, 34(6), 1248-1260.
13. Srivastava, A., & Maheshwari, S. (2018). Some new properties of negation of a probability distribution. International Journal of Intelligent Systems, 33(6), 1133-1145.
14. Sun, C., Li, S., & Deng, Y. (2020). Determining weights in multi-criteria decision making based on negation of probability distribution under uncertain environment. Mathematics, 8(2), 191.
15. Trillas, E. (1979). Sobre funciones de negacion en la teoria de conjuntos difusos, Stochastica 3, 47–59.
16. Xie, D., & Xiao, F. (2019). Negation of basic probability assignment: Trends of Dissimilarity and Dispersion. IEEE Access, 7, 111315-111323.
17. Xie, K., & Xiao, F. (2019). Negation of belief function based on the total uncertainty measure. Entropy, 21(1), 73.
18. Yager, R. R. (2014). On the maximum entropy negation of a probability distribution. IEEE Transactions on Fuzzy Systems, 23(5), 1899-1902.
19. Yin, L., Deng, X., & Deng, Y. (2018). The negation of a basic probability assignment. IEEE Transactions on Fuzzy Systems, 27(1), 135-143.
20. Zhang, J., Liu, R., Zhang, J., & Kang, B. (2020). Extension of Yager's negation of a probability distribution based on Tsallis entropy. International Journal of Intelligent Systems, 35(1), 72-84.